\pdfoutput=1

\documentclass[11pt]{article}

\usepackage[final]{acl}

\usepackage{times}
\usepackage{latexsym}

\usepackage[T1]{fontenc}

\usepackage[utf8]{inputenc}

\usepackage{microtype}

\usepackage{inconsolata}

\usepackage{graphicx}
\usepackage{amsmath}
\usepackage{booktabs}
\usepackage{multirow} 
\usepackage{siunitx} 
\usepackage{pgfmath}
\usepackage{tabularx}

\usepackage{listings}
\usepackage{seqsplit}
\definecolor{tokenMinOneBg}{HTML}{DDF0DD} 
\definecolor{tokenMinTwoBg}{HTML}{E8F5E9} 
\definecolor{tokenMinThreeBg}{HTML}{F0FFF0}
\definecolor{deepgreen}{HTML}{3f701d}

\title{Answer Convergence as a Signal for Early Stopping in Reasoning}

\author{
 \textbf{Xin Liu} \textnormal{\and}
 \textbf{Lu Wang}
\\
Computer Science and Engineering\\
 University of Michigan\\
 Ann Arbor, MI
\\
    \texttt{\{\href{mailto:liuxincs@umich.edu}{liuxincs}, \href{mailto:wangluxy@umich.edu}{wangluxy}\}@umich.edu}
}

\begin{document}
\maketitle
\begin{abstract}
Chain-of-thought (CoT) prompting enhances reasoning in large language models (LLMs) but often leads to verbose and redundant outputs, thus increasing inference cost. We hypothesize that many reasoning steps are unnecessary for producing correct answers. To investigate this, we start with a systematic study to examine what is the minimum reasoning required for a model to reach a stable decision. 
We find that on reasoning tasks like math, models typically converge to their final answers after 60\% of the reasoning steps, suggesting substantial redundancy in the remaining content.
Based on these insights, we propose three inference-time strategies to improve efficiency: (1) early stopping via answer consistency, 
(2) boosting the probability of generating end-of-reasoning signals, and 
(3) a supervised method that learns when to stop based on internal activations.
Experiments across five benchmarks and five open-weights LLMs show that our methods significantly reduce token usage with little or no accuracy drop. 
In particular, on NaturalQuestions, Answer Consistency  reduces tokens by over 40\% while further improving accuracy. 
Our work underscores the importance of cost-effective reasoning methods that operate at inference time, offering practical benefits for real-world applications.\footnote{Additional details about the project are available on its Hugging Face page: \url{https://huggingface.co/spaces/launch/reasoning_earlystop}}
\end{abstract}

\section{Introduction}
Large language models (LLMs) exhibit strong reasoning capabilities through step-by-step generation, known as chain-of-thought (CoT) reasoning \cite{DBLP:conf/nips/Wei0SBIXCLZ22, DBLP:journals/corr/abs-2110-14168, DBLP:conf/aaai/FangD0SCPW24}. However, this approach often leads to unnecessarily long and verbose reasoning traces, resulting in high inference cost and latency. This \textit{overthinking phenomenon} has become a practical bottleneck, especially in real-time or resource-constrained scenarios \cite{DBLP:journals/corr/abs-2503-16419}.

\begin{figure}[t]
    \centering
    \includegraphics[width=0.48\textwidth]{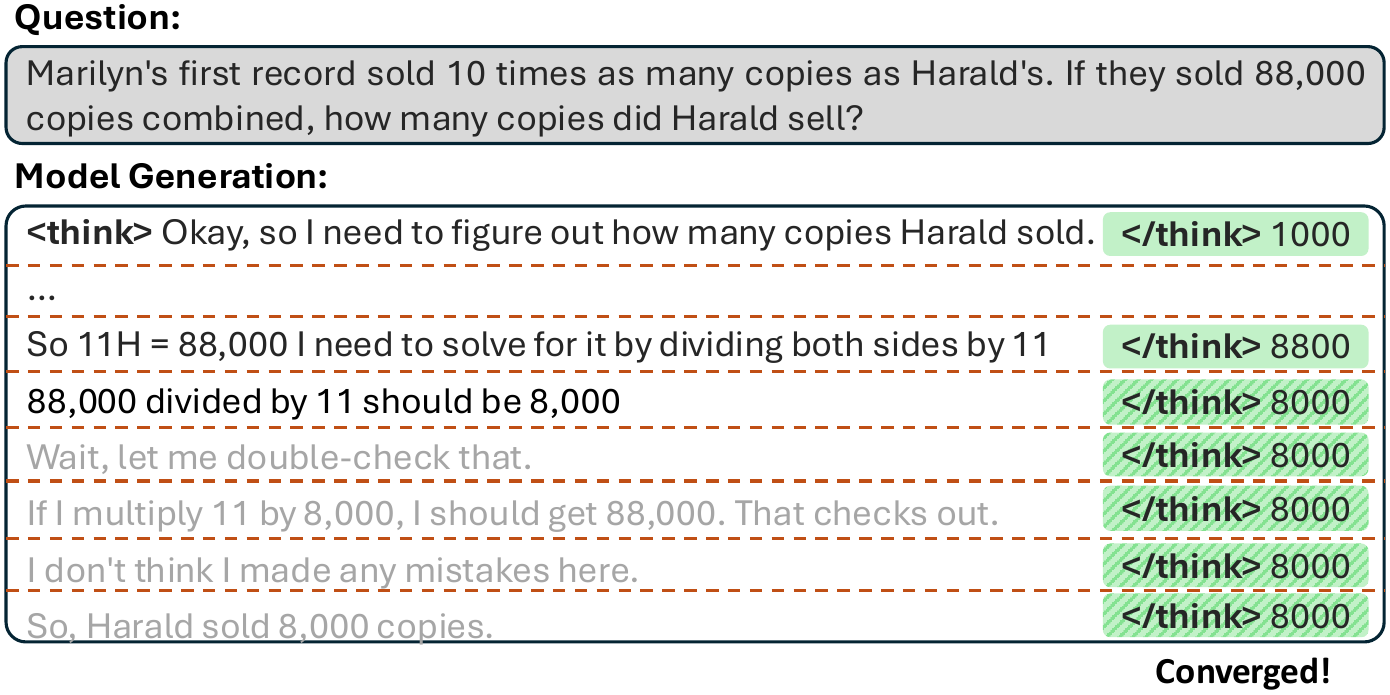}
    \caption{
    R1-Qwen-32B converges early on a GSM8K example, suggesting that later steps could be skipped.
    \vspace{-2mm}
    }
    \label{fig:showcase}
\end{figure}

To improve reasoning efficiency, recent studies have introduced methods that allow models to generate accurate answers with fewer steps. These include reinforcement learning with length-aware rewards \cite{DBLP:journals/corr/abs-2501-12570, hou2025thinkprune}, fine-tuning on variable-length CoT traces \cite{DBLP:journals/corr/abs-2412-18547, DBLP:journals/corr/abs-2502-12067}, and prompt-based approaches that request concise reasoning \cite{DBLP:journals/corr/abs-2502-18600, DBLP:journals/corr/abs-2407-19825, DBLP:journals/corr/abs-2412-18547}. 
These methods typically require retraining on curated data or task-specific prompt design. In contrast, we explore inference-time techniques that improve efficiency without sacrificing accuracy. Particularly, we hypothesize that LLMs often internally converge on an answer before completing the full reasoning trace, an insight we formalize as \textit{answer convergence}. Recognizing such convergence can enable more efficient inference by allowing early stopping without sacrificing accuracy.


To investigate this, we start with a reasoning model early-stopping study that systematically truncates explicit CoTs to assess when the model’s answer generated answer converges, i.e., when the answer remains unchanged despite additional reasoning steps. Our experiments reveal that models often converge well before completing the full reasoning chain, suggesting substantial redundancy and highlighting the potential for improving efficiency through early stopping. Figure~\ref{fig:showcase} shows an example where the model answer converges early despite receiving only partial reasoning, indicating that the remaining steps contribute little to the final prediction. As shown in Figure~\ref{fig:chunking}, this pattern holds across five datasets, even on GPQA, a challenging dataset, many examples converge early.

Motivated by this finding, we propose inference-time strategies to dynamically truncate explicit reasoning based on the observation that models often reach converged answers early: 
(1) Early stopping via \textbf{Answer Consistency}, which halts generation when consecutive reasoning chunks yield identical answers; 
(2) \textbf{Think Token Adjustment}, which encourages models to signal early termination explicitly; and 
(3) a supervised approach, \textbf{Learn-to-Stop}, which utilizes internal activations to predict optimal stopping points. Our methods are \textit{model-agnostic}, require no additional training or LM modification, and significantly reduce inference cost without sacrificing accuracy.
Importantly, this method \textit{does not require ground-truth answer labels}, relying only on self-consistency signals during inference.

We evaluate our methods on five reasoning benchmarks using five open-weights LLMs.
Results show that early stopping strategies substantially reduce token usage.
The supervised approach (\textit{Learn-to-Stop}) consistently maintains accuracy across both simple and complex tasks,
while the two unsupervised methods offer strong efficiency gains particularly on simpler datasets.
Specifically, (1) \textbf{Performance vs.\ token length}: \textit{Learn-to-Stop} cuts up to 48\% of tokens on \textsc{NQ} with \textsc{QwQ-32B}, sometimes even improving accuracy, suggesting that excessive reasoning may introduce unnecessary noise.
(2) \textbf{Supervised vs.\ unsupervised}: Unsupervised methods work well on \textsc{NQ} and \textsc{GSM8K}, while the supervised approach generalizes better to harder tasks like \textsc{MATH-500} and \textsc{GPQA}.

\section{Related Work}

Prior work on improving CoT reasoning efficiency generally falls into four categories. Reinforcement learning-based approaches encourage brevity by introducing length-aware reward functions or adapting reasoning length to problem difficulty~\cite{DBLP:journals/corr/abs-2501-12570, hou2025thinkprune, DBLP:journals/corr/abs-2503-04472}. Supervised fine-tuning methods compress reasoning traces by skipping unimportant tokens~\cite{DBLP:journals/corr/abs-2502-12067}, enforcing token budgets~\cite{DBLP:journals/corr/abs-2412-18547}, or self-training models on shorter rationales through best-of-N sampling and few-shot prompts~\cite{DBLP:journals/corr/abs-2502-20122}. Meanwhile, prompt-based techniques guide models to generate minimal intermediate steps~\cite{DBLP:journals/corr/abs-2502-18600, DBLP:journals/corr/abs-2503-05179}, or use routing mechanisms to dynamically choose efficient reasoning paths or backbones~\cite{DBLP:journals/corr/abs-2501-01306, DBLP:journals/corr/abs-2406-18665}.

In contrast to these methods, which rely on training-time optimization on carefully designed datasets or application-specific prompt design, our work provides lightweight inference-time methods. 
By identifying internal answer convergence, it enables dynamic early stopping without requiring additional training, model modifications, large-scale labeled data, or task-specific prompts.

Concurrent to our research, \citet{DBLP:journals/corr/abs-2504-15895} also address the overthinking problem with a dynamic early exit strategy (DEER). Whereas their method relies on the model's internal confidence in a trial answer , our work proposes complementary approaches based on output consistency and learned stopping signals from activations.

\section{Preliminary}
\label{sec:preliminary}

We investigate whether all steps of a reasoning chain are necessary for the model to converge on its predicted answer, and whether later steps can be omitted without affecting the decision. 
To this end, we first split the CoT into sentence-level chunks, treating each sentence as a distinct reasoning step, 
using the NLTK tokenizer~\cite{bird2009natural}.
These chunks are then incrementally concatenated (e.g., \texttt{chunk1}, \texttt{chunk1+chunk2}, etc.), each followed by an end-of-reasoning token \texttt{</think>}. The model is prompted to generate an answer from each partial chain via greedy decoding. By tracking when the model’s prediction remains unchanged across successive reasoning steps, we identify the earliest point of answer convergence, approximating \textit{the minimum explicit CoT reasoning required for the model to reach a stable decision}.\footnote{We use ``reasoning'' in this paper primarily to refer to explicit chain-of-thought (CoT) reasoning. While models may also perform latent reasoning internally \cite{DBLP:journals/corr/abs-2502-05171, DBLP:journals/corr/abs-2412-06769}, our focus is on improving the efficiency of the explicit reasoning process.}

\begin{figure}[t]
    \centering
    \includegraphics[width=0.43\textwidth]{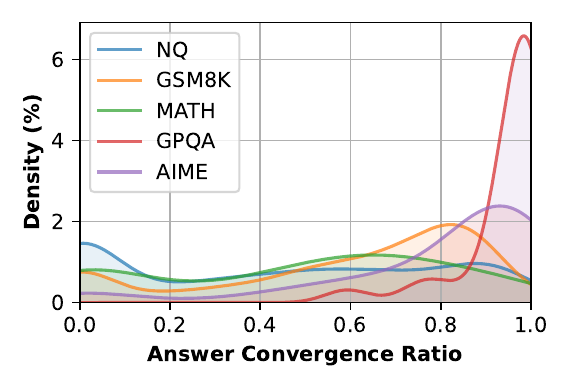}
    \caption{
    Distribution of Answer Convergence Ratios (ACRs) across tasks. Models often converge on their answers early, suggesting that many generated steps might be redundant from the model’s perspective.
    \vspace{-2mm}
    }
    \label{fig:chunking}
\end{figure}

We apply this protocol to five tasks with varying levels of reasoning: \textbf{NaturalQuestions (NQ)}~\cite{DBLP:journals/tacl/KwiatkowskiPRCP19}, \textbf{GSM8K}~\cite{DBLP:journals/corr/abs-2110-14168}, \textbf{MATH-500}~\cite{lightman2023lets}, \textbf{GPQA-Diamond}~\cite{DBLP:journals/corr/abs-2311-12022}, and \textbf{AIME’24}\footnote{\url{https://huggingface.co/datasets/HuggingFaceH4/aime_2024}}. NQ involves minimal reasoning as an information-seeking task, while the others are math and logic benchmarks of increasing difficulty, with AIME’24 containing the most advanced problems. Experiments are conducted using the R1-distilled Qwen-32B model~\cite{deepseekai2025deepseekr1incentivizingreasoningcapability}\footnote{Results for various models are provided in Appendix~\ref{apdx:acr_distribution}.}.
%
To quantify when the model converges, we define the \textbf{Answer Convergence Ratio (ACR)} as the proportion of the explicit reasoning chain required before the predicted answer remains unchanged. Specifically, we first split each CoT trace into sentence-level chunks using the NLTK tokenizer, treating each sentence as one reasoning step. The total number of chunks for a given example defines the original number of steps in the full reasoning chain.
For each instance, we incrementally reveal the reasoning steps and monitor the model's output. We then identify the earliest chunk at which the predicted answer remains consistent through to the end. For example, if the answer stabilizes at the 7th step out of 10 total chunks, the ACR is 0.7.

Figure~\ref{fig:chunking} displays the distribution of ACRs across the five tasks.
We observe that the model often converges early, suggesting that many steps are unnecessary. 
The ACR distribution peaks near 0.0 for NQ, which means nearly no reasoning is needed for the model to come up with its final answer, around 0.8 for GSM8K and MATH-500, and near 0.9 for GPQA and AIME’24, mirroring the increasing reasoning loads. 
These results suggest that early stopping is feasible and can reduce inference-time costs without affecting answer quality.

\section{Early Stopping at Inference Time}
\label{sec:methodology}
We propose three methods to improve reasoning efficiency: two unsupervised approaches based on answer consistency (\S\ref{sec:early_stopping_answer_consistency}) and decoding signals (\S\ref{sec:early_stopping_boosting}) and one supervised method (§\ref{sec:supervised_method}) that predicts when to stop reasoning without retraining the LLM or modifying its parameters.

\subsection{Detecting Answer Consistency}
\label{sec:early_stopping_answer_consistency}
Since LLMs often reach converged answers before completing the full reasoning chain, we introduce an unsupervised stopping criterion based on output consistency. 
During decoding, we monitor the model’s outputs and append the \texttt{</think>} token at predicted natural sentence boundaries, prompting it to produce an answer via greedy decoding.
If the same answer is produced for a fixed number $k$ of consecutive chunks, we consider the reasoning converged and terminate further generation.

\subsection{Think Token Adjustment}
\label{sec:early_stopping_boosting}

During decoding, the model uses the \texttt{</think>} token to indicate the end of reasoning. Ideally, if the model has reached the final answer, it should generate this token early. However, we observe that while \texttt{</think>} often ranks among the top 10 candidates after sentence boundaries, the model tends to prefer tokens like \texttt{wait}, \texttt{or}, or \texttt{but}, which unnecessarily prolong reasoning.

To address this, we boost the probability of \texttt{</think>} during decoding by applying a linear logit transformation\footnote{Implementation details are provided in Appendix~\ref{apdx:boosting_implementation}.}:
\begin{equation}
    y_{t^*} \leftarrow y_{t^*} + \alpha \cdot ( \max(\mathbf{y}) - \frac{1}{|\mathbf{y}|} \sum_i y_i ),
    \label{eq:boosting}
\end{equation}
where $y_{t^*}$ is the logit of the \texttt{</think>} token, $\alpha$ controls the boost strength, $\mathbf{y}$ denotes all vocabulary logits, $y_i$ denotes the logit of $i$-th token in the vocabulary. This encourages the model to terminate reasoning earlier when appropriate.
The added term increases the logit of \texttt{</think>} relative to the average logit, making it more competitive when the model’s output distribution is peaked. 
The term $\max(\mathbf{y}) - \text{mean}(\mathbf{y})$ reflects how peaked the model's output distribution is. This adaptive boost encourages early stopping only when the model is confident, reducing the risk of premature termination.

\begin{figure*}[htbp]
\centering
\centering
\includegraphics[width=1.0\linewidth]{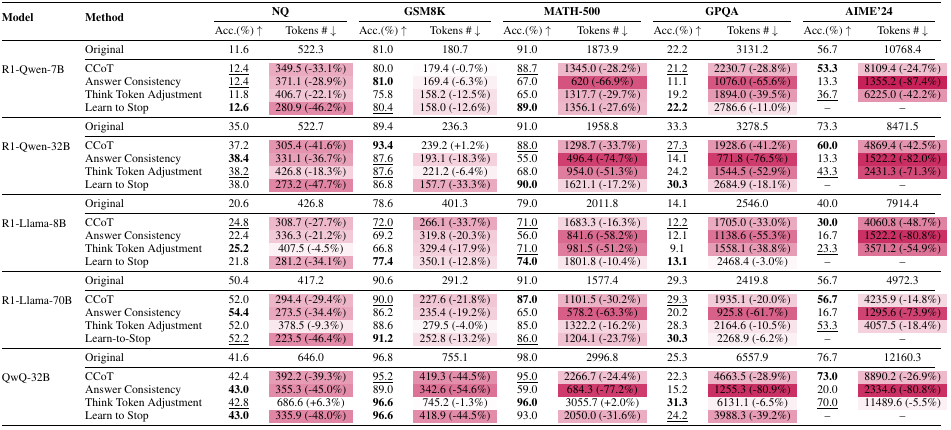}
\caption{
Evaluation results on five reasoning tasks and five model families. For each task, the best and second-best accuracies are shown in \textbf{bold} and \underline{underline}, respectively. In the Tokens \# columns, darker green indicates lower inference cost.
The percentage in parentheses denotes the relative reduction in generated tokens compared to the original model.
Our early stopping methods notably reduce tokens, with Learn-to-Stop offering the best efficiency–performance tradeoff.
For example, on the NQ task, using R1-Qwen-7B and QwQ-32B, the Learn-to-Stop method achieves both the highest accuracy and the largest token reduction among all baselines, illustrating its strong efficiency–performance tradeoff.
\vspace{-2mm}
}
\label{tab:model_performance}
\end{figure*}

\subsection{Learning When to Stop Reasoning}
\label{sec:supervised_method}
Recent work shows that LLM activations encode useful signals such as knowledge and confidence \cite{DBLP:conf/nips/KapoorGRCPBWDGW24, DBLP:conf/emnlp/LiuBW24}. We hypothesize that they also capture reasoning progress, including when to stop. Specifically, the final-layer activation \( \mathbf{h}_t \) may reflect both the model’s certainty and the need for further computation.


To leverage this, we train a supervised model to predict optimal stopping points using the model’s internal activations. Given the sequential nature of reasoning, we use an LSTM to encode the activation sequence \( \{ \mathbf{h}_1, \dots, \mathbf{h}_T \} \). At each chunk \( t \), the LSTM output \( \mathbf{z}_t \) is passed to a sigmoid classifier: \( \hat{p}_t = \sigma(\mathbf{W} \mathbf{z}_t + b) \), where \( \hat{p}_t \in [0, 1] \) represents the probability of stopping at step \( t \).

Training labels are constructed by identifying the earliest chunk where the predicted answer matches the final answer and remains unchanged.
Note that our method does not rely on gold-standard answers, enabling unsupervised training based solely on the model’s self-predicted outputs.
Chunks from that point onward are labeled 1, and all earlier chunks are labeled 0.
We optimize a binary cross-entropy loss:
\begin{equation}
\mathcal{L} = -\frac{1}{T} \sum_{t=1}^{T} \left[ p_t \log \hat{p}_t + (1 - p_t) \log (1 - \hat{p}_t) \right]
\end{equation}
where \( p_t \in \{0, 1\} \) is the ground truth label, and \( T \) is the number of chunks. At inference time, reasoning stops when \( \hat{p}_t \geq \tau \), with threshold \( \tau \) tuned on validation data. This approach enables us to utilize the model’s internal dynamics to \textit{improve reasoning efficiency without training or modifying the base LLM}.

\section{Experimental Setup}
\label{sec:experiments}

We evaluate our methods on the five tasks introduced in \S\ref{sec:preliminary}, with dataset splits and statistics detailed in Appendix~\ref{apdx:dataset_statistics}.
Experiments are conducted on five LLMs across three families: R1-distilled Qwen and Llama~\cite{deepseekai2025deepseekr1incentivizingreasoningcapability}, and QwQ~\cite{qwq32b}.
We report Accuracy \textbf{(Acc.)}, average generated tokens (\textbf{Tokens \#}), and token reduction (\%) relative to the original model.
As baselines, we include the original model without length control and Concise CoT (CCoT)~\cite{DBLP:journals/corr/abs-2407-19825}, which constrains reasoning length using prompt token budgets. We follow their setup by setting a fixed token budget.\footnote{The implementation details and prompts are provided in Appendix~\ref{apdx:implementation_details}. }

\begin{table}[t]
    \centering
    \small
    \renewcommand{\arraystretch}{0.93}
    \begin{tabularx}{\linewidth}{X}
    \toprule
    \textbf{Question:} Carlos is planting a lemon tree. The tree will cost \$90 to plant. Each year it will grow 7 lemons, which he can sell for \$1.5 each. It costs \$3 a year to water and feed the tree. How many years will it take before he starts earning money on the lemon tree? \\ \hline
    \textbf{Original}: To determine how many years 
    ......
    Solving for n: n = \$90 / \$7.5, n=\colorbox{green}{12} \\
    \textcolor{gray}{Wait, let me double-check my calculations to make sure I didn't make a mistake ... That seems correct.}\\
    </think>\verb|\boxed|\{12\} \\
    \textbf{Token \#}: 439\\
    \hline
    \textbf{Learn to Stop:} Okay, so Carlos is planting 
    ......
    Calculating that, 90 divided by 7.5 equals \colorbox{green}{12}. \\ 
    </think>\verb|\boxed|\{12\}\\
    \textbf{Token \#}:338 \\ \hline
    \textbf{Model:} R1-Qwen-32B \\
    \bottomrule
    \end{tabularx}
    \caption{
        Example from GSM8K comparing the original model and Learn-to-Stop. The original generates \textcolor{gray}{redundant reasoning} after reaching \colorbox{green}{12}, while Learn-to-Stop terminates reasoning for a more concise response.
        \vspace{-2mm}
    }
    \label{tab:case_study}
\end{table}

\section{Results and Analysis}
\label{sec:results}

From the results in Table~\ref{tab:model_performance}, we make the following key findings:

\paragraph{Early stopping further improves performance on tasks with low reasoning demand.} On NQ, which involves minimal reasoning, all early stopping methods match or surpass the original model’s accuracy while greatly reducing token usage. This suggests that the original model may overthink and generate unnecessary reasoning steps, which even harms the performance. 
Manual inspection reveals that the original model often overthinks, such as adding unnecessary self-reflection or even revising correct answers, whereas early stopping prevents such degradation.\footnote{Examples in Appendix~\ref{apdx:case_study_NQ}.}
The early stopping methods effectively mitigate this issue by terminating reasoning earlier, leading to improved performance.
%

\paragraph{Unsupervised methods are effective for simpler tasks.} On NQ and GSM8K, Answer Consistency and Think Token Adjustment reduce tokens without harming accuracy, despite requiring no additional training. However, on harder tasks like MATH-500 and GPQA, their performance becomes less stable, indicating that shallow signals like output consistency may be insufficient under high reasoning complexity.\footnote{The ablation study of answer consistency and think token adjustment is provided in Appendix \ref{apdx:ablation}.}

\paragraph{Learn-to-Stop offers robust efficiency gains.}  This supervised method consistently balances accuracy and token savings across models and tasks. For example, on GSM8K with QwQ-32B, it reduces tokens by 44\% (755.1 to 418.9) with only a 0.2\% drop in accuracy; on the harder task GPQA, it cuts 39.2\% tokens with comparable performance (24.2\% vs. 25.3\%). Compared to CCoT, which explicitly enforces a token budget at the prompt level, Learn-to-Stop achieves greater reductions while maintaining or improving accuracy. 
This advantage stems from its ability to leverage rich internal model signals (e.g., hidden activations) to determine optimal stopping points.
Unlike methods that rely on annotated answers or external supervision, Learn-to-Stop infers training labels by detecting when the model’s predicted answer stabilizes, enabling fully unsupervised training.
As a result, it generalizes well across tasks with varying reasoning complexity.

Since Learn-to-Stop operates purely at inference time and keeps the prompt unchanged, it is compatible with prompt-level strategies like CCoT, offering opportunities for further gains through combination. Table~\ref{tab:case_study} exemplify how Learn-to-Stop avoids redundant reasoning while preserving output quality.\footnote{Examples of other datasets in Appendix~\ref{apdx:case_study}.}

\section{Effect of Model Size}
\label{sec:model-size}

Our study covers five LLMs of varying sizes across three families (Qwen, Llama, QwQ).
As shown in Table~\ref{tab:model_performance} and Appendix Figure~\ref{fig:chunk_dist_minipage},
we observe two consistent trends.

\paragraph{Convergence Dynamics.}
Larger models stabilize on answers earlier, with ACR distributions shifting leftward. For example, R1-Llama-70B requires fewer reasoning steps to converge compared to its 8B variant on \textsc{NQ} and \textsc{GSM8K}.
This is likely because their enhanced capacity enables them to synthesize information and resolve problem components more efficiently internally, reducing their reliance on generating an extended, explicit chain of thought to reach a stable conclusion.

\paragraph{Efficiency Gains.}
Token savings also improve with scale.
Learn-to-Stop reduces tokens by $46.2\%$ on \textsc{NQ} for R1-Qwen-7B,
versus $47.7\%$ for R1-Qwen-32B.
Similarly, R1-Llama-70B achieves a $46.4\%$ reduction compared to $34.1\%$ for 8B.
This suggests that stronger models converge more confidently and thus benefit more
from early stopping.

\section{Conclusion}

We study how to reduce redundancy in chain-of-thought (CoT) reasoning to improve LLM inference efficiency. Across five benchmarks and five open-weight LLMs, we find that answer convergence often occurs early, revealing substantial redundancy. Based on this, we propose three inference-time methods that stop generation once reasoning is sufficient. These methods cut token usage by up to 40\% without accuracy loss, offering a practical alternative to full-chain reasoning without retraining or model changes.

\section*{Acknowledgments}
This work is supported in part by National Science Foundation through grant 2046016. We thank anonymous reviewers of ARR for their thoughtful comments. 

\section*{Limitations}
While our proposed early stopping strategies significantly reduce inference cost with minimal or no loss in accuracy, several limitations remain. First, our methods rely on the assumption that the model’s answer convergence correlates with the correctness of the final output. However, convergence does not guarantee correctness, especially in tasks with higher reasoning complexity (e.g., GPQA and AIME).
Second, by enabling models to make predictions without observing the full reasoning trace, our approach may compromise the faithfulness of reasoning.
A manual analysis suggests that most truncated traces remain aligned with the final answer, though occasional unfaithful stops do occur (see Appendix~\ref{apdx:faithfulness} for details).
Future work should aim to jointly optimize for both faithfulness and conciseness, ensuring that reasoning remains both efficient and trustworthy.

\bibliography{custom}

\appendix

\section{Appendix}
\label{sec:appendix}

\subsection{ACR Distribution on Various Models}
\label{apdx:acr_distribution}

\begin{figure*}[htbp]
    \centering

    \begin{minipage}[b]{0.45\textwidth}
        \centering
        \includegraphics[width=\linewidth,page=1]{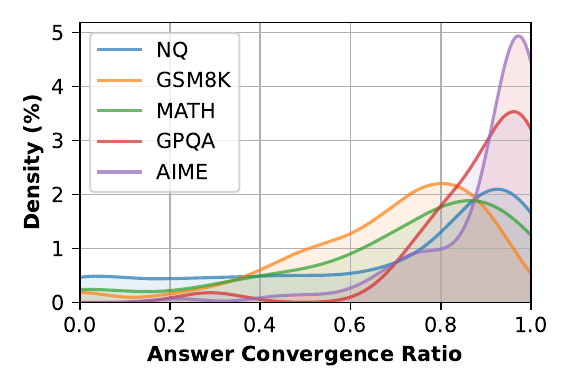}
        \caption*{(a) R1-distilled-Llama-8B}
    \end{minipage}
    \hfill
    \begin{minipage}[b]{0.45\textwidth}
        \centering
        \includegraphics[width=\linewidth,page=1]{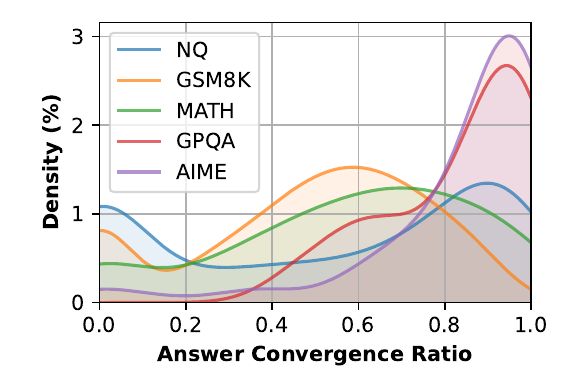}
        \caption*{(b) R1-distilled-Llama-70B}
    \end{minipage}
    
    \vspace{5mm}
    
    \begin{minipage}[b]{0.45\textwidth}
        \centering
        \includegraphics[width=\linewidth,page=1]{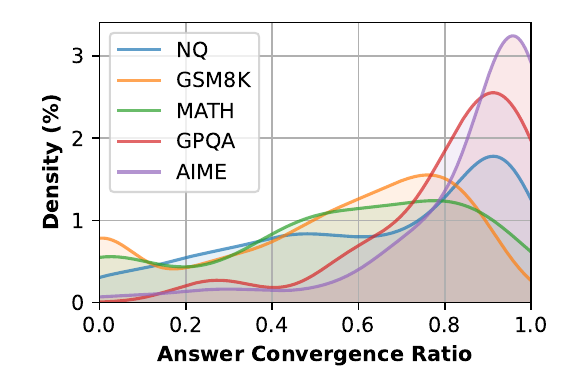}
        \caption*{(c) R1-distilled-Qwen-7B}
    \end{minipage}
    \hfill
    \begin{minipage}[b]{0.45\textwidth}
        \centering
        \includegraphics[width=\linewidth,page=1]{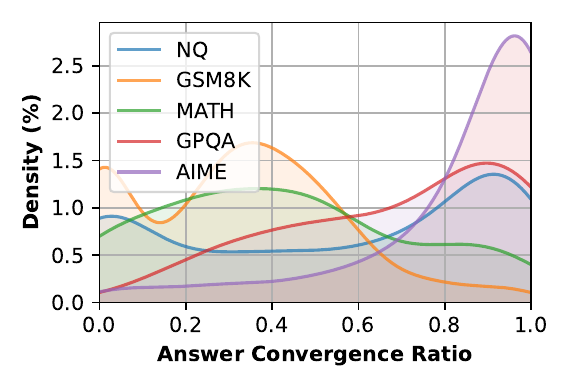}
        \caption*{(d) QwQ-32B}
    \end{minipage}

    \caption{ACR distributions of various models.}
    \label{fig:chunk_dist_minipage}
\end{figure*}

We present the ACR distributions for several models: R1-distilled Llama-8B, R1-distilled Llama-70B, R1-distilled Qwen-7B, and QwQ-32B. The results are shown in Figure~\ref{fig:chunk_dist_minipage}. All models exhibit similar trends to the R1-distilled Qwen-32B baseline, confirming their ability to produce stable answers before completing the full reasoning chain. Moreover, tasks with higher reasoning demands tend to correspond to higher ACRs. When comparing distributions across models, we observe that larger models generally achieve lower ACRs, suggesting that they require fewer reasoning steps to converge on an answer. This implies that larger models may possess a more efficient internal reasoning process, enabling them to reach final answers more quickly.

\subsection{Dataset Splits and Statistics}
\label{apdx:dataset_statistics}
For tasks with available training data (NQ and GSM8K), we sample 1,000 examples for training the supervised method and 100 for validation. For tasks without predefined training data (MATH-500 and GPQA-Diamond), we reserve 100 examples as the test set and split the remaining data into 80\% for training and 20\% for validation. Due to the limited size of AIME'24, which includes only 30 test examples, we evaluate only the unsupervised methods on this task.
The statistics for each task are shown in Table~\ref{tab:dataset_statistics}.
\begin{table}[t]
\centering
\begin{tabular}{lccc}
\toprule
\textbf{Tasks} & \textbf{Train} & \textbf{Validation} & \textbf{Test} \\
\midrule
NQ                        & 800  & 200  & 3610 \\
GSM8K                     & 800  & 200  & 1319 \\
MATH-500                  & 320  & 80   & 100  \\
GPQA-Diamond              & 78   & 20   & 100  \\
AIME'24                 & --   & --   & 30   \\
\bottomrule
\end{tabular}
\caption{Dataset statistics of each task.}
\label{tab:dataset_statistics}
\end{table}

\subsection{Implementation of Early Stopping via Boosting End of Think Token}
\label{apdx:boosting_implementation}
To implement boosting of the \texttt{</think>} token, we design a logit processor that adjusts the model’s output logits before sampling. Specifically, it increases the logit of the \texttt{</think>} token according to \S\ref{sec:early_stopping_boosting}. Once the \texttt{</think>} is generated, the processor is disabled to prevent further modifications. This mechanism allows the model to emit the \texttt{</think>} token earlier if it has already reached a confident answer. We integrate the logit processor into the VLLM framework~\cite{DBLP:conf/sosp/KwonLZ0ZY0ZS23}, enabling efficient logit manipulation during decoding without compromising throughput.
We run all experiments three times and report the average results.

\subsection{Implementation Details}
\label{apdx:implementation_details}
When applying early stopping via answer consistency, we empirically set the number of consecutive chunks $k$ to 10. For early stopping via boosting, we set the hyperparameter \(\alpha\) to 0.6. Regarding the supervised method, we use a single-layer LSTM with 128 hidden units and a dropout rate of 0.1. The model is trained for 200 epochs with a batch size of 32, using the Adam optimizer with a learning rate of \(5e^{-4}\). The confidence threshold \(\tau\) is tuned on the validation set, and we set it to 0.50, 0.99, 0.99, and 0.50 for NQ, GSM8K, MATH-500, and GPQA-Diamond, respectively. For the CCoT baseline, we set the token budget to 100 for all tasks. We run all experiments three times and report the averaged results.

We use the VLLM framework \cite{DBLP:conf/sosp/KwonLZ0ZY0ZS23} to sample the model outputs for all experiments to ensure efficient inference. All the prompts we used are provided in Appendix~\ref{apdx:prompts}.

\subsection{Ablation Study}
\label{apdx:ablation}

\noindent\textbf{Setup.}
We ablate two key hyperparameters used by our unsupervised methods:
(i) $\alpha$ for \emph{Think Token Adjustment}, which up-weights the probability of the end-of-reasoning marker; and
(ii) $k$ for \emph{Answer Consistency}, the consecutive-consistency threshold.
Unless otherwise noted, we report results with R1-distilled-Llama3.1-8B across five benchmarks
(\textsc{NQ}, \textsc{GSM8K}, \textsc{MATH-500}, \textsc{GPQA}, \textsc{AIME'24}).

\subsubsection{Effect of $\alpha$ (Think Token Adjustment)}

\paragraph{Accuracy.}
Moderate boosting ($\alpha \le 0.4$) maintains accuracy on easier tasks while larger $\alpha$ can cause premature termination and degrade performance on harder reasoning tasks.

\begin{table}[h]
\centering
\small
\setlength{\tabcolsep}{5.2pt}
\begin{tabular}{lccccc}
\toprule
$\alpha$ & NQ & GSM8K & MATH-500 & GPQA & AIME'24 \\
\midrule
0.0 & 20.6 & 78.6 & 79.0 & 14.1 & 40.0 \\
0.2 & 23.2 & 74.6 & 79.0 & 14.1 & 40.0 \\
0.4 & 24.6 & 73.2 & 73.0 & 14.1 & 40.0 \\
0.6 & 25.2 & 66.8 & 71.0 & 9.1 & 23.3 \\
0.8 & 20.4 & 31.2 & 50.0 & 8.1 & 30.0 \\
1.0 & 15.8 & 43.2 & 67.0 & 8.1 & 30.0 \\
\bottomrule
\end{tabular}
\caption{Effect of $\alpha$ on \textbf{accuracy} (\%).}
\label{tab:alpha-acc}
\end{table}

\paragraph{Token Usage.}
As $\alpha$ increases, tokens drop sharply; however, overly large $\alpha$ harms accuracy despite aggressive savings.

\begin{table}[h]
\centering
\small
\setlength{\tabcolsep}{5.2pt}
\begin{tabular}{lccccc}
\toprule
$\alpha$ & NQ & GSM8K & MATH-500 & GPQA & AIME'24 \\
\midrule
0.0 & 426.8 & 401.3 & 2011.8 & 2546.0 & 7914.4 \\
0.2 & 426.8 & 366.4 & 2011.8 & 2546.0 & 7914.4 \\
0.4 & 418.9 & 349.2 & 1712.5 & 2393.9 & 5060.0 \\
0.6 & 418.5 & 316.1 & 1017.9 & 1557.0 & 2429.8 \\
0.8 & \ \,91.3 & \ \,49.1 & \ \,68.5 & \ \,79.1 & \ \,81.4 \\
1.0 & \ \,17.6 & \ \,10.3 & \ \,11.2 & \ \,27.8 & \ \,9.3 \\
\bottomrule
\end{tabular}
\caption{Effect of $\alpha$ on \textbf{\#tokens}.}
\label{tab:alpha-tok}
\end{table}

\noindent\textit{Takeaway.} $\alpha\!\in\![0.2,0.4]$ offers a good accuracy–efficiency trade-off; aggressive boosting ($\alpha\!\ge\!0.6$) risks early truncation on hard tasks.

\subsubsection{Effect of $k$ (Answer Consistency)}

\paragraph{Accuracy.}
Small $k$ stops too early and hurts accuracy; performance improves and saturates around $k\!\in\![20,30]$.

\begin{table}[h]
\centering
\small
\setlength{\tabcolsep}{5.2pt}
\begin{tabular}{lccccc}
\toprule
$k$ & NQ & GSM8K & MATH-500 & GPQA & AIME'24 \\
\midrule
\ \,2 & 22.4 & 15.6 & 14.0 & 3.0 & 0.0 \\
\ \,5 & 22.8 & 46.4 & 35.0 & 5.1 & 0.0 \\
10 & 22.2 & 69.0 & 56.0 & 12.1 & 16.7 \\
15 & 22.2 & 76.6 & 63.0 & 10.1 & 23.3 \\
20 & 21.8 & 77.8 & 69.0 & 11.1 & 26.7 \\
25 & 22.0 & 78.0 & 72.0 & 10.1 & 26.7 \\
30 & 21.4 & 78.4 & 74.0 & 11.1 & 26.7 \\
\bottomrule
\end{tabular}
\caption{Effect of $k$ on \textbf{accuracy} (\%).}
\label{tab:k-acc}
\end{table}

\paragraph{Token Usage.}
Larger $k$ delays stopping and increases tokens; the curve flattens beyond $k\!\approx\!25\!-\!30$.

\begin{table}[h]
\centering
\small
\setlength{\tabcolsep}{5.2pt}
\begin{tabular}{lccccc}
\toprule
$k$ & NQ & GSM8K & MATH-500 & GPQA & AIME'24 \\
\midrule
\ \,2 & \ \,77.2 & \ \,65.6 & \ \,86.1 & 105.8 & \ \,84.4 \\
\ \,5 & 222.3 & 204.1 & 418.6 & 464.2 & 414.9 \\
10 & 336.3 & 319.8 & 841.6 & 1138.6 & 1522.2 \\
15 & 381.8 & 364.7 & 1177.1 & 1463.2 & 2507.7 \\
20 & 402.0 & 379.1 & 1585.3 & 1754.0 & 3558.5 \\
25 & 412.7 & 385.2 & 1676.5 & 1949.6 & 4247.2 \\
30 & 420.1 & 396.1 & 1750.3 & 2098.8 & 4445.9 \\
\bottomrule
\end{tabular}
\caption{Effect of $k$ on \textbf{\#tokens}.}
\label{tab:k-tok}
\end{table}

\noindent\textit{Takeaway.}
$k\!\in\![20,30]$ yields near-saturated accuracy but lower efficiency; practical settings can choose $k$ around $10\!-\!20$ to balance accuracy and token savings.

\subsection{Effect of Early Stopping on Faithfulness}
\label{apdx:faithfulness}

Early stopping enables models to generate predictions without completing the full reasoning trace, which raises concerns about the faithfulness of the generated rationale.
Faithfulness, i.e., whether the reasoning chain logically supports the final answer, has been a key focus in recent studies~\cite{DBLP:journals/corr/abs-2307-13702, chen2025reasoning}.
To assess this aspect, we conduct a manual evaluation of reasoning chains generated by R1-Qwen-32B on GPQA under the Learn-to-Stop setting. 
We restrict our analysis to instances where the original model produced a correct answer and categorize the early-stopped generations into the following four classes:

\begin{itemize}
    \item \textbf{Perfect Stop}: The model halts reasoning immediately after predicting the correct answer for the first time, and the preceding steps form a faithful chain of reasoning.
    \item \textbf{Late Stop}: The model already reaches the correct answer but continues with unnecessary reasoning, suggesting it could have stopped earlier.
    \item \textbf{Incorrect Stop}: The model halts reasoning too soon, leading to an incorrect answer.
    \item \textbf{Unfaithful but Correct Stop}: The model outputs the correct answer, but the truncated reasoning trace does not support the answer logically. This is often caused by stopping before completing a faithful CoT.
\end{itemize}

\begin{figure}[t]
    \centering
    \includegraphics[width=0.95\linewidth]{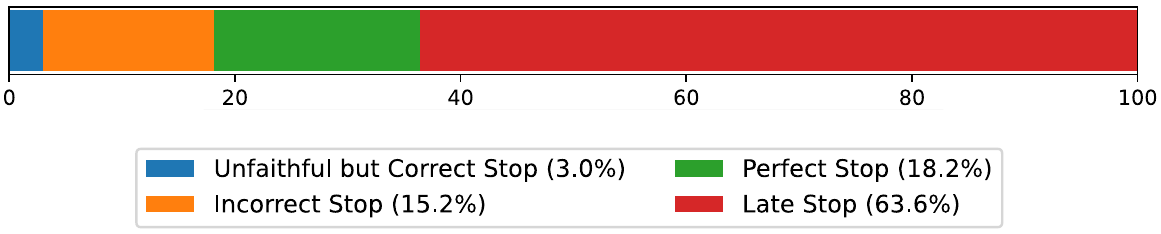}
    \caption{Faithfulness analysis of the Learn-to-Stop method on GPQA.}
    \label{fig:faithfulness_analysis}
\end{figure}

We present the results in Figure~\ref{fig:faithfulness_analysis}. The majority of cases fall into the \textbf{Stop Late} category (63.6\%), indicating that while the model reaches the correct answer, it continues with unnecessary reasoning, suggesting a suboptimal stopping point.
This highlights the opportunity to further refine stopping criteria by incorporating finer-grained signals, e.g. hidden states from more layers that more precisely detect convergence and faithfulness in real time.
The \textbf{Stop Perfectly} cases account for 18.2\%, demonstrating the model's ability to identify the correct point to stop.
Only 3\% of cases are labeled as \textbf{Unfaithful}, where the correct answer is produced without supporting reasoning. This suggests that although our method may occasionally lead to unfaithful outputs, such instances are relatively rare in \textsc{GPQA}. Overall, these results indicate that our approach can reduce reasoning steps without significantly compromising the faithfulness of the generated reasoning.
Future work could explore hybrid strategies that combine answer stability with faithfulness-aware signals, enabling models to adaptively balance efficiency and reasoning integrity across diverse tasks.

\subsection{Test Cases from NQ}
\label{apdx:case_study_NQ}
We list two examples from NQ in Table~\ref{tab:case_study_NQ}. In the first example, the original model initially generates the correct answer (Moira Kelly), but then adds unnecessary reasoning steps, including self-reflection, and revises the answer to an incorrect one (Julie Kavner). Similarly, in the second example, the model starts with the correct answer (Ashoka), but again adds unnecessary reasoning, ultimately revising it to an incorrect answer (Kalinga Nanda).

\begin{table*}[t]
    \centering
    \small
    \begin{tabularx}{\linewidth}{X}
    \toprule
    \textbf{Question:} who does the voice of nala in the lion king \\ \hline
    \textbf{Original}: Okay, so I need to figure out who voices Nala in The Lion King. I remember watching the movie when I was a kid, but I can't recall the voice actress. Let me think. I know that in the original 1994 version, some of the main voices were well-known actors. Simba was voiced by Jonathan Taylor Thomas, Mufasa by James Earl Jones, and Scar by Jeremy Irons. But Nala... I'm not sure. Maybe it was someone like Jennifer Hudson? Wait, no, she was in the 2019 remake, I think. In the original, I think it was a different actress. Maybe someone like, I don't know, \textcolor{deepgreen}{Moira Kelly}? No, that doesn't sound right. Maybe it's someone else. Let me think of other voice actors from that era. Maybe it's someone like \textcolor{red}{Eliza Dushku}? Or maybe it's someone less famous. I think it's a British actress. Oh, maybe it's someone like, I don't know, \textcolor{red}{Julie Kavner}? Wait, no, she did Mufasa's sister in the original... \\
    \textbf{Correct Answer \#}: \textcolor{deepgreen}{Moira Kelly}\\
    \hline
    \textbf{Question:} the king of kalinga in the mauryan period was \\ \hline
    \textbf{Original}: Okay, so I need to figure out who the king of Kalinga was during the Mauryan period. Hmm, I remember that the Mauryan Empire was a significant empire in ancient India, and it was around the 3rd century BCE. The most famous king from that period is \textcolor{deepgreen}{Ashoka} the Great, known for his role in the Kalinga war.\textbackslash n \textbackslash n Wait, but the question is about the king of Kalinga, not the Mauryan king. So \textcolor{deepgreen}{Ashoka} was the Mauryan emperor, and he conquered Kalinga. I think the king of Kalinga at that time might have been a local ruler who resisted \textcolor{deepgreen}{Ashoka}'s expansion. There's a specific name that comes to mind, something like \textcolor{red}{Kalinga Nanda}. I'm not entirely sure, but I recall that Nanda was a common name for kings in that region, and \textcolor{red}{Kalinga Nanda} was the ruler who opposed \textcolor{deepgreen}{Ashoka}.\textbackslash n \textbackslash n I should check if there are any other possible names. Some sources mention that the king of Kalinga was indeed \textcolor{red}{Kalinga Nanda}, and that \textcolor{deepgreen}{Ashoka}'s conquest of Kalinga was a major event in his reign, leading to his conversion to Buddhism. So yeah, \textcolor{red}{Kalinga Nanda} is the name I'm looking for... \\
    \textbf{Correct Answer \#}: \textcolor{deepgreen}{Ashoka}\\
    \textbf{Model:} R1-Qwen-32B \\
    \bottomrule
    \end{tabularx}
    \caption{
        Examples from NQ with responses from the original model.
    }
    \label{tab:case_study_NQ}
\end{table*}

\subsubsection{Case Study}
\label{apdx:case_study}
We provide a case study of the original model and the Learn-to-Stop method in Table~\ref{tab:case_study_apdx}. 

\begin{table*}[t]
    \centering
    \small
    \begin{tabularx}{\linewidth}{X}
    \toprule
    \textbf{Question:} Each week, between 30 and 50 students show up for an archery class run by Betty and Wilma.  Usually the students break up into groups of equal size for target practice.  However, this week, Betty noticed that she could not break the students up into multiple groups of equal size.  Wilma noticed that if she and Betty both joined the students in practicing, they still could not break the archers up into groups of equal size.  How many students showed up to the archery class this week? \\ \hline
    \textbf{Original}: Okay, so I have this math problem here about Betty and Wilma's archery class. Let me try to understand what it's asking......47:\textbackslash n 47 + 2 = 49. 49 is not prime because 7*7=49. So 47 is out.\textbackslash n \textbackslash nSo the only number that fits both conditions is \colorbox{green}{41}. \textcolor{gray}{When you add 2, it becomes 43, which is also prime. Therefore, there must have been 41 students at the archery class this week.}\\
    </think>\verb|\boxed|\{41\} \\
    \textbf{Token \#}: 741\\
    \hline
    \textbf{Learn-to-Stop:} Okay, so I need to figure out how many students showed up to the archery class this week. The problem says that each week, between 30 and
    ......
    \colorbox{green}{41} fits both conditions. \\ 
    </think>\verb|\boxed|\{41\}\\
    \textbf{Token \#}:673 \\ \hline
    \textbf{Dataset:} MATH-500 \\
    \hline
    \textbf{Question:} In a specific region of the sky, astronomers have observed that the number of stars varies with parallax as 1/plx\^{}{}5. How does the number of stars in that region of the sky change with distance (per unit range of distance, r)? \\ \hline
    \textbf{Original}: Okay, so I've got this problem here where astronomers observed that the number of stars varies with parallax as 1 over plx to the fifth power. Hmm, I need to figure out how the number of stars changes with distance, specifically per unit range of distance r...... Therefore, 5k / (1/r\^{}6 * r) = 5k r\^{}6 / r = 5k r\^{}4. Yes, that's correct. So, the number of stars per unit distance is proportional to \colorbox{green}{r\^{}4}.
    \textcolor{gray}{Wait, but that would mean that as you go further out, the number of stars per unit distance increases rapidly. That seems odd because usually, the number of stars in a volume element increases with the square of the radius (since volume is 4dr), but here it's increasing faster than that. Maybe because the number density itself is increasing with distance? Wait, no. The problem states that the number of stars varies with parallax as 1/plx\^{}5. So, if parallax decreases (distance increases), the number increases as 1/plx\^{}5, which is equivalent to d\^{}5. So, the total number up to distance d is proportional to d\^{}5. But the number per unit distance would require considering the derivative. Wait, maybe I'm confusing cumulative counts with differential counts. Let me clarify. If N(pl) is the total number of stars with parallax less than pl, then N(pl) = k / pl\^{}5. But parallax less than pl corresponds to distances greater than d = 1/pl. So, N(d) = k d\^{}5. But then the number of stars beyond distance d would be N\_total - N(d) = k d\^{}5. Hmm, that doesn't make much sense because as d increases, N(d) increases, meaning more stars are at larger distances, which is expected. But to find the number per unit distance, we take the derivative of N(d) with respect to d. So, dN/dd = 5k d\^{}4. So, the number of stars per unit distance at distance d is proportional to d\^{}4. Wait, but that would imply that the number density is increasing with distance, which might not align with standard stellar distributions unless there's an actual increase in star density with distance, which isn't typically the case. Normally, the number density of stars in the galaxy decreases with distance from the galactic center, but in this problem, we're probably considering a region where the distribution follows this specific law. So, putting it all together, the number of stars per unit distance (r) is proportional to r\^{}4.}\\
    </think>\verb|\boxed|\{r\^{}4\} \\
    \textbf{Token \#}: 1952 \\
    \hline
    \textbf{Learn to Stop:} Okay, so I have this problem where the number of stars varies with parallax as 1 over plx to the fifth power. Hmm, I remember that parallax is related to distance. Let me think about how they're connected.\textbackslash n \textbackslash n Wait, parallax (pl) is inversely proportional to distance (d)
    ......
    Therefore, the number of stars per unit distance (N(r)) is proportional to \colorbox{green}{r\^{}4}.\\ 
    </think>\verb|\boxed|\{r\^{}4\}\\
    \textbf{Token \#}:1431 \\ \hline
    \textbf{Dataset:} GPQA-Diamond \\
    \bottomrule
    \end{tabularx}
    \caption{
        Examples from MATH-500 and GPQA-Diamond with responses from the original model and the Learn-to-Stop method. The original model generates redundant reasoning steps after producing the correct answer (41 and r\^{}4), while the Learn-to-Stop method terminates reasoning earlier, resulting in a more concise response.
    }
    \label{tab:case_study_apdx}
\end{table*}

\subsection{Prompts}
\label{apdx:prompts}

\subsubsection{NQ}
\label{apdx:prompts_nq}
{\tiny\begin{lstlisting}
Answer the following question.
Directly output your final answer within \\boxed{}. DO NOT say anything else.
 

Question: {question}
\end{lstlisting}
}

\subsubsection{Math Reasoning Tasks}
\label{apdx:prompts_math}
{\tiny\begin{lstlisting}
Solve the following math problem. 
Directly output your final answer within \\boxed{}. DO NOT say anything else.
 

Question: {question}
\end{lstlisting}
}

\subsubsection{CCoT Prompt}
\label{apdx:prompts_ccot}
{\tiny\begin{lstlisting}
Answer the following question.
You should think step by step, and limit the thinking process length to {LENGHT_BUGGED} words.
Directly output your final answer within \\boxed{}. DO NOT say anything else.
 

Question: {question}
\end{lstlisting}
}

{\tiny\begin{lstlisting}
Solve the following math problem. 
You should think step by step, and limit the thinking process length to {LENGHT_BUGGED} words.
Directly output your final answer within \\boxed{}. DO NOT say anything else.
 

Question: {question}
\end{lstlisting}
}

\end{document}